\documentclass[psfig,twocolumn]{IEEEtran}
\usepackage{graphicx}
\usepackage{amsmath}
\usepackage{array}
\usepackage{multicol}
\newfont{\Bb}{msbm10}
\newtheorem{theorem}{\bf Theorem}
\begin{document}

%\title{On the Influence of Sparsity-Aware Nodes on the Performance of Sparse Diffusion Heterogeneous Networks}
\date{}
\title{Sparse Distributed Learning via Heterogeneous Diffusion Adaptive Networks}
\date{}
\author{Bijit K. Das$^{1}$, Mrityunjoy Chakraborty$^{2}$, and Jer\'{o}nimo Arenas-Garc\'{i}a$^{3}$\\
   $^{1,\;2}$ Department of Electronics and Electrical Communication Engineering,\\
  Indian Institute of Technology, Kharagpur, INDIA.\\
  $^{3}$ Department of Signal Theory and Communications,\\
   Univ. Carlos III de Madrid, Spain.\\
  E.Mail : $^1$bijitbijit@gmail.com, $^2$mrityun@ece.iitkgp.ernet.in,  $^3$jarenas@tsc.uc3m.es}

%\author{%
%\authorblockN{%
%Bijit Kumar Das\authorrefmark{1} and Mrityunjoy
%Chakraborty\authorrefmark{2} }
%%
%\authorblockA{%
%\authorrefmark{1}
%Indian Institute of Technology, Kharagpur 721302 West Bengal India \\
%E-mail: bijitbijit@gmail.com  Tel: +91-9051271617}
%%
%\authorblockA{%
%\authorrefmark{2}
%Indian Institute of Technology, Kharagpur 721302 West Bengal India \\
%E-mail: mrityun@ece.iitkgp.ernet.in   Tel: +91-3222-283512}
%%
%}

%\author{
%\authorblockN{Bijit Kumar Das}
%\authorblockA{
%Indian Institute of Technology\\
%Kharagpur 721302 West Bengal India \\
%E-mail: bijitbijit@gmail.com\\
%Tel: +91-9051271617}
%\and
%\authorblockN{Mrityunjoy Chakraborty}
%\authorblockA{
%Indian Institute of Technology\\
%Kharagpur 721302 West Bengal India\\
%E-mail: mrityun@ece.iitkgp.ernet.in\\
%Tel: +91-3222-283512}
%}

\maketitle
\thispagestyle{empty}

%\begin{multicols}{2}
\begin{abstract}
 In-network distributed estimation of sparse parameter vectors via diffusion LMS strategies has been studied and investigated
 in recent years. In all the existing works, some convex regularization approach has been used at each node of the network in order to
 achieve an overall network performance superior to that of the simple diffusion LMS, albeit at the cost of increased computational overhead.
 In this paper, we provide analytical as well as experimental results which show that the convex regularization can be selectively applied
 only to some chosen nodes keeping rest of the nodes sparsity agnostic, while still enjoying the same optimum behavior as can be realized by deploying
 the convex regularization at all the nodes. Due to the incorporation of unregularized learning at a subset of nodes, less computational cost is needed in the proposed approach.
 We also provide a guideline for selection of the sparsity aware nodes and  a closed form expression for the optimum regularization parameter.
 %The claims made are strongly supported by simulation studies.

 \end{abstract}
 {\bf Index terms}--Adaptive network, diffusion LMS, Sparse systems, excess mean
 square error, adaptive filter, $l_1$ norm.

\section{Introduction}

%
% In distributed estimation literature, the well-known decentralized strategies like the average consensus and the gossip algorithms \cite{Scaglione}-\cite{Barbarossa} have been extensively studied to address several
% control engineering problems.

% In general, these algorithms require two different time-scales. One scale is to take samples of the observable quantities across the nodes, and another time-scale is for using those
% measurements in an iterative fashion to reach an agreement. In a real-time scenario where the measurable data keeps flowing into the processing nodes these conventional approaches
% are not exactly suitable.

Diffusion strategies \cite{Sayed1}-\cite{Sayed3} were first
invented to solve distributed estimation problems in real-time
environments where data are continuously streamed. Here, all nodes
employ adaptive filter algorithms to process the streaming data,
and simultaneously share their instantaneous estimates with their
neighbors. These approaches are also very useful to model many
self-organizing systems \cite{Sayed4}.

Recently, in \cite{Zhang}-\cite{Sayed5}, diffusion LMS schemes
have been used to estimate sparse vectors, or equivalently, to
identify FIR systems that have most of the impulse response
coefficients either zero or negligibly small. In these papers,
certain sparsity promoting norms of the filter coefficient vectors
have been used to regularize the standard LMS cost function,
prominent amongst them being the $l_1$ norm of the coefficient
vector that leads to the sparsity aware, zero attracting LMS
(ZA-LMS) \cite{Gu}-\cite{Bijit2} form of weight adaptation. These
diffusion sparse LMS algorithms manifest superior performance in
terms of lesser steady state network mean square deviation (NMSD)
compared with the simple diffusion LMS.
%These approaches have been studied and investigated in numerous works in the field of stand-alone adaptive filters \cite{Gu}-\cite{Bijit2}.
% Sparse adaptive filters are useful in a number of application areas \cite{Radecka}-\cite{Dutt} such as acoustic echo cancellation,
% network echo cancellation, sparse channel estimation etc.

In this paper, we show that the minimum level of the steady state
NMSD achieved using ZA-LMS based update at \textit{all} the nodes
of the network can also be obtained by a \textit{heterogeneous}
network with only a fraction of the nodes using the ZA-LMS update
rule (referred as sparsity aware nodes in this paper) while the
rest employing the standard LMS update (referred as sparsity
agnostic nodes in this paper), provided the nodes using the ZA-LMS
are distributed over the network maintaining some ``uniformity".
Note that reduction in the number of sparsity aware nodes reduces
the overall computational burden of the network, especially when
more complicated sparsity aware algorithms involving significant
amount of computation are deployed to exploit sparsity. As shown
in this paper, the only adjustment to be made to achieve the above
reduction in the number of sparsity aware nodes is a proportional
increase in the value of the optimum zero attracting coefficient.
Analytical expressions explaining the above behavior are provided
and the claims made are validated via detailed simulation studies.
Finally, the proposed analysis, though restricted to the
$l_1$-norm regularized algorithm (i.e., ZA-LMS) only, can be
trivially extended to the case of more general norms and thus
similar behavior can also be expected from the corresponding
heterogeneous networks.
%that arise in different application areas like target localization, environment monitoring and many sensor network problems.

\section{Brief Review of Diffusion Sparse LMS Algorithms}

%In the co-operative mode of operation, the nodes interact with their neighbors by sharing information.

We consider a connected network consisting of $N$ nodes that are
spatially distributed.  At every time index $n$, each $k^{th}$
node collects some scalar measurement $d_k(n)$ and some $M \times
1$ vector ${\bf u}_k(n)$ which are related by the following model:
\begin{equation}
 d_k(n) = {\bf u}_{k}^{T}(n){\bf w}_0 + v_k(n),
\end{equation}
where $v_k(n)$ is the measurement noise at the $k^{th}$ node and
${\bf w}_0$ is the unknown $M \times 1$ vector, known a priori to
be sparse, which is required to be estimated. Both ${\bf u}_k(n)$
and $v_k(n)$ are variates generated from some Gaussian
distributions, with ${\bf u}_k(n)$ and $v_k(m)$ being mutually
independent for all $n,\;m$.

In the diffusion scheme, every $k$-th node, $k\,=\,1,\,
2,\cdots,\,N$ deploys a $M \times 1$ adaptive filter ${\bf
w}_k(n)$ to estimate ${\bf w}_0$, which takes $d_k(n)$ and ${\bf
u}_k(n)$ respectively as the local desired response and input
vectors. The estimates of ${\bf w}_0$, i.e., ${\bf w}_k(n)$ for
each $k$ are exchanged with the neighbors of the $k$-th node,
i.e., nodes directly connected to it, and are used to refine the
estimates in one of the two following manners : (A)
Adapt-then-Combine (ATC) where ${\bf w}_k(n)$ is first updated to
an intermediate estimate ${\bf v}_k(n+1)$, which is then linearly
combined with similar estimates received from the neighbors, and
(B) Combine-then-Adapt (CTA) where ${\bf w}_k(n)$ is first
linearly combined with similar estimates received from the
neighbors and then updated. Originally, the diffusion schemes were
proposed assuming LMS form of weight adaptation at each node
\cite{Sayed2}-\cite{Sayed3}. In the context of sparse estimation,
certain sparsity exploiting norms of ${\bf w}_k(n)$ were added to
the corresponding LMS cost function \cite{Zhang}-\cite{Sayed5},
the most popular of them being the $l_1$ norm penalty $||{\bf
w}_k(n)||_1$ which results in the introduction of the zero
attracting terms $sgn[{\bf w}_k(n)]$ in the LMS update equations
\cite{Gu}-\cite{Mei}. The resulting diffusion ZA-LMS algorithm for
the ATC scheme, popularly termed as ZA-ATC diffusion algorithm
\cite{Sayed5}, is shown in Table I and is considered by us in this
paper.
%
%exchanges and fuses its local estimate with the estimates of its
%neighbors i.e., the nodes directly connected to it. Let $\aleph_k$
%denotes the set of nodes in the neighborhood of the node $k$
%including itself.
%
%Each $i^{th}$ node works like an adaptive filter ${\bf w}_i(n)$ which uses the pair ($d_i(n), {\bf u}_i(n)$).
%
%Due to the recent advents in the popular fields of compressing sensing and sparse signal processing in general, different
%families of sparse adaptive filters have been proposed and investigated. One of the most popular approaches is $l_q$ norm regularized
%LMS algorithms where $0\leq q\leq 1.$
%In \cite{Zhang} and \cite{Sayed5}, the stand-alone sparse adaptive filtering mechanisms have been extended to distributed case.
%In \cite{Sayed5}, sparsity inducing update terms have been used only in adaptation step of a diffusion LMS algorithm.
%In \cite{Zhang}, slightly different steps have been incorporated.

The parameter $\rho$ in Table I is the zero-attracting coefficient
which is a very very small, positive constant taken same for all
the nodes and $\aleph_k$ denotes the set of nodes in the
neighborhood of the node $k$ (including itself).
% in \cite{{Sayed5}} and also in this paper.

% \fbox{
% \addtolength{\linewidth}{-2\fboxsep}%
% \addtolength{\linewidth}{-2\fboxrule}%
% \begin{minipage}[t][1.2\height][c]{0.9\linewidth}
%  \begin{eqnarray}
%     e_k(n) &=& d_k(n) - {\bf w}_{k}^{T}(n){\bf u}_k(n)\nonumber\\
%    {\bf v}_k(n+1) &=& {\bf w}_k(n) + \mu_k {\bf u}_k(n)e_k(n)\nonumber\\ &-& \rho sgn[{\bf w}_k(n)]\\
%    {\bf w}_k(n+1) &=& \sum_{j\in \aleph_k} c_{j,k}^{'}{\bf v}_j(n+1)
%  \end{eqnarray}
% \end{minipage}
%}

\begin{table}[ht]
\caption{The ZA-ATC Diffusion Algorithm \cite{Sayed5}} \fbox{
 \addtolength{\linewidth}{-2\fboxsep}%
 \addtolength{\linewidth}{-2\fboxrule}%
 \begin{minipage}[t][1.2\height][c]{0.9\linewidth}
  \begin{eqnarray}
     e_k(n) &=& d_k(n) - {\bf w}_{k}^{T}(n){\bf u}_k(n)\nonumber\\
    {\bf v}_k(n+1) &=& {\bf w}_k(n) + \mu_k {\bf u}_k(n)e_k(n)\nonumber\\ &-& \rho sgn[{\bf w}_k(n)]\\
    {\bf w}_k(n+1) &=& \sum_{j\in \aleph_k} c_{j,k}^{'}{\bf v}_j(n+1)
  \end{eqnarray}
  \end{minipage}
}
  \label{table:za_atc}
\end{table}

 The combining coefficients $c_{l,k}^{'}$ are non-negative constants which are usually chosen
 satisfying the following \cite{Sayed1} :
\begin{eqnarray}
 c_{l,k}^{'} &>& 0  \hspace{1mm} \textit{if}  \hspace{1mm}  l \in \aleph_k \nonumber\\
         &=& 0 \hspace{1mm} \textit{elsewhere}. \nonumber\\
 \textit{and} \hspace{1mm} \sum_{l \in \aleph_k} c_{l,k}^{'} &=&
 1.
\end{eqnarray}
There exist several standard schemes in the literature to choose
the coefficients $c_{i,j}^{'}$, e.g., the uniform combination
rule, the metropolis rule, the Laplacian rule and the nearest
neighbor rule to name a few. Using these coefficients, a
combination matrix ${\bf C}^{'}$ is defined for the network, where
$[{\bf C}^{'}]_{i,j}=c_{i,j}^{'}.$

\section{Proposed Heterogeneous Network and its NMSD Behavior}
Before presenting the proposed heterogeneous network and its at
par behavior with the ZA-ATC based diffusion network of
\cite{Sayed5}\footnote{The networks presented in \cite{Sayed5} and
also in \cite{Zhang} are ``homogeneous" in the sense that these
networks deploy only sparsity aware nodes.}, it will be useful to
consider some of the major results of \cite{Sayed5} here. For
this, we first define the average network mean-square deviation at
the $n^{th}$ time index as,
\begin{eqnarray}
 MSD_{net}(n) = \frac{1}{N}\sum_{k=1}^{N}MSD_k(n),
\end{eqnarray}
where $MSD_k(n)$ is the individual mean-square deviation of the
$k^{th}$ node at the $n^{th}$ time index, i.e.,
$$ MSD_k(n) = E[\|{\bf w}_0 -  {\bf w}_k(n)\|^2] = E[\|{\bf {\tilde w}}_k(n)\|^2]$$
where ${\bf {\tilde w}}_k(n)={\bf w}_0 -  {\bf w}_k(n)$ is the
weight deviation vector for the $k$-th node at $n$-th index.

 The expression for steady-state $MSD_{net}(n)$ (i.e., $MSD_{net}(\infty)$) of the ZA-ATC algorithm
 was derived analytically in \cite{Sayed5}. However, \cite{Sayed5}
 considered a more general form of diffusion, in which
 both $d_l(n)$ and ${\bf u}_l(n)$ are also exchanged with the neighbors
 along with the local estimates ${\bf w}_l(n)$.
 In contrast, in this paper, we consider exchange of only ${\bf
 w}_l(n)$ which is also the most common form of diffusion.
 Additionally, we introduce a few more simplifications in
 \cite{Sayed5}. Firstly, we assume same step-size $\mu$ for all
 nodes. Next, both the input signal and
noise at each node are assumed to be spatially and temporally
i.i.d. Under these, it is easy to check that the
$MSD_{net}(\infty)$ expression for the ZA-ATC algorithm
\cite{Sayed5} simplifies to the following :
\begin{eqnarray}
 MSD_{net}(\infty) &=& \frac{\mu^2\sigma_{v}^{2}\sigma_{u}^{2}}{N}[vec({\bf C}^T{\bf C})]^T({\bf I} - {\bf F})^{-1}{\bf q} \nonumber\\
                   &+&  \frac{1}{N}(\beta(\infty) - \alpha(\infty)),
                   \label{eq:MSD}
\end{eqnarray}

with
\begin{equation}
 \alpha(\infty) = -2\mu E[sgn[{\bf w}(\infty)]^T{\boldsymbol \varOmega}{\bf C}{\bf C}^T({\bf I} - \mu{\bf D}){\bf {\tilde w}}(\infty)]
\end{equation}
and
\begin{equation}
 \beta(\infty) = \mu^2 E[\|sgn[{\bf w}(\infty)]\|_{{\boldsymbol \varOmega}{\bf C}{\bf C}^T{\boldsymbol
 \varOmega}}^{2}],
\end{equation}
where $vec(\cdot)$ is an operator that stacks the columns of its
argument matrix on top of each other, ${\bf q} = vec({\bf I}_{MN
\times MN})$, ${\bf w}(n) = col({\bf w}_1(n), \hspace{1mm} {\bf
w}_2(n), \cdots {\bf w}_N(n)) $ and ${\bf {\tilde w}}(n) =
col({\bf {\tilde w}}_1(n), \hspace{1mm} {\bf {\tilde w}}_2(n),
\dots {\bf {\tilde w}}_N(n))$, with
$col(\cdot,\cdot,\cdots,\cdot)$ denoting an operator that carries
out stacking of its argument column vectors on top of each other,
and $\sigma_{v}^{2}$ and $\sigma_{u}^{2}$ are the variances of the
noise and input signal respectively. The matrices $\bf C$, ${\bf
D}$, ${\bf F}$ and ${\boldsymbol \varOmega}$ are defined as follows :\\
${\bf C} = {\bf C}^{'}\otimes {\bf I}_{M \times M}$
[$\otimes$ defines the right Kronecker product.]\\
${\bf D} = \sigma_{u}^{2}{\bf I}_{MN \times MN},$\\
${\bf F} = (1 - 2 \mu \sigma_{u}^{2} + \mu^2 \sigma_{u}^{4}) ({\bf C}\otimes {\bf C})$,\\
${\boldsymbol \varOmega} = \rho{\bf I}$. [Also note that for a
vector $\bf a$ and a matrix $\bf B$, $||{\bf a}||_{\bf B}^2$
indicates ${\bf a}^T{\bf B}{\bf a}$.]

It is noticed that the first term in the R.H.S. of (6) is actually
the steady-state network MSD of simple ATC diffusion LMS
\cite{Sayed2} and is independent of $\rho$. Let us denote the
second term as $\phi(\rho)$, i.e., $ \phi(\rho) =
\frac{1}{N}(\beta(\infty) - \alpha(\infty))$. It is easy to see
that one can express $\phi(\rho)$ as
$\phi(\rho)=-\alpha^{'}(\infty)\rho+\beta^{'}(\infty){\rho}^2$,
where,
$$\alpha^{'}(\infty) = -2\mu E[sgn[{\bf w}(\infty)]^T{\bf C}{\bf C}^T({\bf I} - \mu{\bf D}){\bf {\tilde w}}(\infty)]$$
and
$$ \beta^{'}(\infty) = \mu^2E[\|sgn[{\bf w}(\infty)]\|_{{\bf C}{\bf C}^T}^{2}]~(>\;0).$$
The function $\phi(\rho)$ has two zero-crossing points, one at
$\rho=0$ and the other at
$\rho=\frac{\alpha^{'}(\infty)}{\beta^{'}(\infty)}$, and between
them, $\phi(\rho)$ takes only negative values with the minima
occurring at $\rho=\frac{\alpha^{'}(\infty)}{2\beta^{'}(\infty)}$,
which, from (6), also minimizes $MSD_{net}(\infty)$. For systems
that are highly sparse, it follows from \cite{Gu} that
$\alpha^{'}(\infty)>0$ and conversely, for non-sparse systems,
$\alpha^{'}(\infty)<0$. Since for proper zero attraction, $\rho$
must be positive, the optimum value of $\rho$ is then given by
\begin{equation}
 \rho_{opt} = max[0,
 \frac{\alpha^{'}(\infty)}{2\beta^{'}(\infty)}].
\end{equation}
The corresponding minimum value of $\phi$ (when $\rho_{opt}>0$) is
then given as
\begin{equation}
 \phi_{min} = -\frac{{\alpha^{'}}^2(\infty)}{4N\beta^{'}(\infty)}.
\end{equation}
\textbf{The Proposed Heterogeneous Diffusion Network
:}\\\\
In this section, we show that the same level of $\phi_{min}$ as
given by (10) and therefore, the same $min[MSD_{net}(\infty)]$ can
be reached by a heterogeneous network as well, where only a
fraction of the nodes are sparsity aware and rest are sparsity
agnostic, provided the network is designed satisfying the
assumptions I.A and I.B as given in the box below where $S$
denotes the set of indices of the sparsity aware nodes and :

\fbox{
 \addtolength{\linewidth}{-2\fboxsep}%
 \addtolength{\linewidth}{-2\fboxrule}%
 \begin{minipage}[t][1.2\height][c]{0.9\linewidth}
\begin{center}
Assumption I\\%$ Assumption \,  I$\\
\end{center}

I.A\\
We assume that the matrix ${\bf C}^{'}$ is doubly stochastic,
i.e., $\forall i,j,$
$\sum_{i=1}^{N} c_{i,j}^{'} = 1$ and $\sum_{j=1}^{N} c_{i,j}^{'} = 1.$ This is valid for many practical rules used to select combiner coefficients.\cite{Sayed1}\\

I.B\\
We also assume that the sparsity-aware nodes are distributed over
the network in such a way that $\forall j,$ $\sum_{i \in
S}c_{i,j}^{'} = \frac{N_s}{N}$.
% the $k^{th}$ node is connected to approximately $\lfloor\frac{N_s N_k}{N}\rfloor$ sparsity-aware nodes
% including itself ($N_k$ is the degree of the neighborhood $k^{th}$ node has), $ \sum_{i \in S}{\bf c}_{i}^{'} \approx \frac{N_s}{N}{\bf 1}_{N\times 1}.$
% Here, we also use the doubly stochastic property of ${\bf C}^{'},$ and assume that the non-zero combiner weights in each column (and in each row) have almost same value.\
.\\\\
\emph{The physical interpretation of this assumption is that it
ensures a uniform influence of the sparsity aware nodes on each
node of the network. This can be employed as a design criterion.}
     \end{minipage}
}\\\\

In order to show the above, we replace  the matrix ${\boldsymbol
\varOmega}$ by a new one defined
 as ${\boldsymbol \varOmega}_s = diag[\rho_1{\bf I}_{M \times M}, \rho_2{\bf I}_{M \times M},
 \dots \rho_k{\bf I}_{M \times M} \dots \rho_{N}{\bf I}_{M \times M} ],$
 where $$\rho_k = \rho, \hspace{2mm} if \hspace{2mm} k \in S,
 \hspace{2mm} else \hspace{2mm} \rho_k = 0$$.

Using this and the fact that ${\bf I}-\mu{\bf
D}=(1-\mu\sigma_u^2){\bf I}$, $\alpha(\infty)$ and $\beta(\infty)$
modify to $\alpha_1(\infty)$ and $\beta_1(\infty)$, given as
follows :
\begin{eqnarray}
 & & \alpha_{1}(\infty)\nonumber\\
 &=& -2\mu(1-\mu\sigma_u^2)E[sgn[{\bf w}(\infty)]^T{\boldsymbol \varOmega}_s{\bf C}{\bf C}^{T}{\bf {\tilde w}}(\infty)]
\end{eqnarray}
and
\begin{eqnarray}
 & & \beta_{1}(\infty)\nonumber\\
 &=& \mu^2 E[sgn[{\bf w}(\infty)]^T{\boldsymbol \varOmega}_s{\bf C}{\bf C}^{T}{\boldsymbol \varOmega}_s sgn[{\bf w}(\infty)]]
 \end{eqnarray}
Note that unlike $\alpha(\infty)$ and $\beta(\infty)$, it is lot
more difficult to express $\alpha_1(\infty)$ and $\beta_1(\infty)$
as a function of $\rho$, since unlike ${\boldsymbol \varOmega}$,
${\boldsymbol \varOmega}_s$ can not be written simply as $\rho{\bf
I}$. Instead, one needs to analyze the RHS of (11) and (12) to
express $\alpha_1(\infty)$ and $\beta_1(\infty)$ in terms of
$\rho$. Towards this, we make the following assumptions :\\ \fbox{
 \addtolength{\linewidth}{-2\fboxsep}%
 \addtolength{\linewidth}{-2\fboxrule}%
 \begin{minipage}[t][1.2\height][c]{0.9\linewidth}
\begin{center}
 Assumption II\\
\end{center}

II.A\\
$ E[sgn[{\bf w}_i(\infty)]{\bf {\tilde w}}_{m}^{T}(\infty)]
\approx {\boldsymbol \theta},$ ($\forall i,m$)  \
is a matrix independent of $i$ and $m$, when $m \in \aleph_j$, $\forall j \in \aleph_i$\\

II.B\\
$ E[sgn[{\bf w}_i(\infty)]sgn[{\bf w}_{m}^{T}(\infty)]] \approx
{\boldsymbol \psi},$ ($\forall i,m$)   \ is a matrix independent
of $i$ and $m$, when $m \in \aleph_j$, $\forall j \in
\aleph_i$.\\\\
\emph{In words, the above assumptions tell that, at the
steady-state, any pair of nodes having overlapping neighborhood
(including directly connected nodes, and the same node) show
approximately same cross-covariance and similar cross-moments.
This is motivated by the fact that all nodes have same step-size,
same input and noise statistics, and the abovementioned pairs
continuously exchange their intermediate estimates using diffusion
strategy.}
\end{minipage}
}

It is then possible to prove the following :\\
\begin{theorem}
For a network satisfying the $Assumptions~{\bf I.A}$ and ${\bf
II.A}$ as given above, we have,
\begin{eqnarray}
 \alpha_{1}(\infty) = -2\rho \mu (1-\mu\sigma_u^2)Tr[{\boldsymbol \theta}] N_s.
\end{eqnarray}
Proof: Skipped due to page limitation.\\
\end{theorem}
%Here, we also use the doubly stochastic property of ${\bf C}^{'},$ and assume that the non-zero combiner weights in each column (and in each row) have almost same value.\\

\begin{theorem}
For a network satisfying the $Assumptions~ {\bf I.A}$, ${\bf I.B}$
and ${\bf II.B}$ as given above, we have,
\begin{eqnarray}
 \beta_{1}(\infty) =  \frac{\mu^2\rho^{2}Tr[{\boldsymbol
 \psi}]N_{s}^{2}}{N}.
\end{eqnarray}
Proof: Skipped due to page limitation.\\
\end{theorem}

\begin{figure*}[ht]
\begin{center}
\includegraphics[width=200mm, height=65mm]{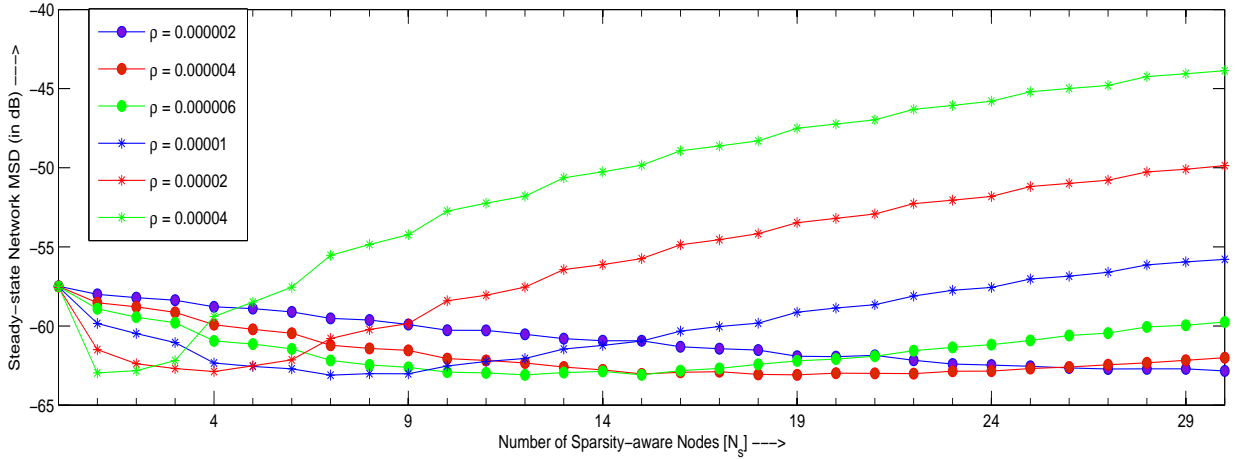}
%\includegraphics[width=175mm]{lnl.eps}
%\fbox{ \rule[-b17mm]{0pt}{30mm} FIGURE }
\end{center}
\caption{The Network MSD versus number of sparsity-aware nodes
($N_s$) curves for different values of $\rho$}
\vspace*{-3pt}%{\hfill\footnotesize The symmetric property of the
%coefficients reduces the length of the coefficient vector by
%half.\hfill}
\end{figure*}

Substituting $\alpha_{1}(\infty)$ and $\beta_{1}(\infty)$ in $
\phi(\rho) = \frac{1}{N}(\beta_1(\infty) - \alpha_1(\infty))$,
then differentiating w.r.t. $\rho$ and equating the derivative to
zero, we obtain,
\begin{equation}
 \rho_{opt} = max[0, -\frac{(1-\mu\sigma_u^2)Tr[{\boldsymbol \theta}]N}{\mu Tr[{\boldsymbol
 \psi}]N_s}].
\end{equation}

The corresponding minimum value of $\phi(\rho)$ [when
$\rho_{opt}>0$, i.e., the system is sparse], say, $\phi_{min}'$ is
given as
\begin{equation}
 \phi_{min}' = -\frac{(1-\mu\sigma_u^2)^2Tr[{\boldsymbol \theta}]^2}{Tr[{\boldsymbol
 \psi}]}.
\end{equation}
Note that $\phi_{min}'$ as given in (16) is independent of $N_s$.
Therefore, \textit{its value remains same when $N_s=N$, i.e., when
the network becomes} \textit{homogeneous with all nodes being
sparsity aware}. This also implies that if $\phi_{min}$ as given
by (10) is analyzed using the assumptions I and II, it would give
rise to the same expression as that of $\phi_{min}'$ (i.e., (16)).
From this and (15), we then make the following two
conclusions :\\\\
$\bullet$ The $min[MSD_{net}(\infty)]$ does not change when the
network changes from being homogeneous to heterogeneous, with only
$N_s$ of the total $N$ ($0< N_s\le N$)nodes employing sparsity
aware adaptation.\\
$\bullet$ For sparse systems, the $\rho_{opt}$ minimizing
$\phi(\rho)$ and thus $[MSD_{net}(\infty)]$ (i.e.,
$-\frac{(1-\mu\sigma_u^2)Tr[{\boldsymbol \theta}]N}{\mu
Tr[{\boldsymbol \psi}]N_s}$ as given in (15)) is inversely
proportional to $N_s$, meaning that while maintaining the same
$min[MSD_{net}(\infty)]$, one can reduce the number of sparsity
aware nodes by introducing proportional increase in the value of
$\rho$.

 \section{Simulation Studies}
To test the performance of the heterogeneous networks, we use a
strongly connected network of $N = 30$ nodes placed randomly in a
geographic region. The weights of the edges are determined by the
uniform combination rule \cite{Sayed1}. The goal of the network is
to estimate a $128 \times 1$ vetor ${\bf w}_0$ which is highly
sparse (only one coefficient being non-zero). We choose the same
step-size $\mu = 6 \times 10^{-3}$ for all the nodes. Among these
$30$ nodes, $N_s$ number of nodes use the ZA-LMS and rest of the
nodes use simple LMS update, with the former spaced 'uniformly'
(i.e., satisfying assumptions I.A and I.B) over the network. The
input signals and noise variables are drawn from Gaussian
distributions, and they are temporally and spatially independent.
Also, the input and noise statistics are same for all the nodes,
with $\sigma_{u}^{2} = 1$, and $\sigma_{v}^{2} = 1 \times
10^{-4}$. To start with, the value of $\rho$ is kept fixed at $2
\times 10^{-6}$ for all the $N_s$ sparsity aware nodes. The
simulation is then carried out for $3000$ iterations and the
network steady state MSD is evaluated by taking ensemble average
over $1000$ independent runs. This is done for different values of
$N_s$ (ranging from $0$ to $30$) and based on this, the network
steady state MSD is plotted as a function of $N_s$. The value of
$\rho$ is then increased progressively to take the following five
values : $4 \times 10^{-6}, 6 \times 10^{-6}, 1 \times 10^{-5}, 2
\times 10^{-5}, 4 \times 10^{-5}$, one at a time for all the
ZA-LMS based nodes. Fig. 1 displays the network steady state MSD
vs. $N_s$ plots with $\rho$ as a parameter. It is easily seen from
Fig. 1 that (i) the minima reached by each MSD-vs-$N_s$ plot is
same for all the plots, and (ii) as $\rho$ increases, the value of
$N_s$ where the minima occurs reduces and vice versa. In other
words, Fig. 1 validates the theoretical conjectures made in the
previous section.

 %as predicted by the theoretical
%analysis of the previous section, different values of optimum
%$\rho$ are required to achieve the \textit{same} minimum network
%MSD for different number of sparsity-aware nodes $N_s$.

%
%
%
%\section{Conclusions}
%In this paper, we try to show that we can estimate a sparse vector using a diffusion strategy which employs
%sparsity-aware but computationally intensive algorithms only at a subset of all the nodes.
%This approach is mathematically proved to be as efficient as the conventional diffusion sparse algorithms \cite{Zhang}, \cite{Sayed5},
%but requires less computational burden. For wireless sensor networks, this can be seen as a new mechanism for hardware-friendly and efficient real-time
%distributed estimation of sparse vectors.
%

%\end{multicols}

%\begin{multicols}{1}

\end{document}